\documentclass{article}

\usepackage{arxiv}
\usepackage[utf8]{inputenc} 
\usepackage[T1]{fontenc}    
\usepackage{hyperref}       
\usepackage{url}            
\usepackage{booktabs}       
\usepackage{amsfonts}       
\usepackage{amsmath}
\usepackage{nicefrac}       
\usepackage{microtype}      
\usepackage{lipsum}
\usepackage{graphicx}
\usepackage{listings}
\usepackage{float}
\usepackage{pgfplots}
\pgfplotsset{compat=1.18}
\usepgfplotslibrary{groupplots}

\usepackage[
  backend=biber,
  style=numeric,
  sorting=none,
  maxbibnames=2,
  minbibnames=1,
  maxcitenames=2,
  mincitenames=1
]{biblatex}

\title{DMF: A Deterministic Memory Framework for Conversational AI Agents}

\author{
 Matteo Stabile \\
  Roma Tre University, Italy\\
  mat.stabile@stud.uniroma3.it\\
   \And
 Enrico Zimuel\thanks{Corresponding author} \\
  Roma Tre University, Italy\\
  enrico.zimuel@uniroma3.it\\
}

\begin{document}
\maketitle
\begin{abstract}
Conversational AI agents require memory systems that are both scalable and semantically coherent across long interaction horizons.
Existing approaches rely predominantly on large language model (LLM)-based summarisation at write time, which introduces non-determinism, escalating token costs, and opacity in pruning decisions.
We present the \textbf{Deterministic Memory Framework (DMF)}, a CPU-first approach that replaces generative memory compression with a fully deterministic pipeline grounded in classical NLP analysis, vector geometry, and mathematical scoring.
DMF assigns each conversational interaction a \emph{Survival Score} $\Omega$ computed from deterministic content signals, conversational cues, and structured provenance, combined through a logistic projection.
An \emph{interaction-count decay law}, denoted as $\Omega_{\mathrm{eff}}(\Delta n)$, governs how relevance evolves as new turns arrive, where $\Delta n$ is the number of newer interactions rather than wall-clock time, preserving full determinism.
We present the mathematical formulation of DMF, its structured recall pipeline, the pruning decision procedure, and the evaluation protocol. Experiments are conducted on a purpose-built benchmark using the \textit{LoCoMo} and \textit{LongMemEval} datasets. We compare DMF against \textit{Mem0}, a popular memory layer for AI agents. DMF achieves comparable accuracy while using zero tokens to prepare the memory context and 5× to 242× fewer tokens over the entire conversation. These results show that it is possible to eliminate LLM calls from the memory-management loop, reducing token costs to nearly zero and enabling deterministic memory systems for conversational AI agents.
\end{abstract}

\section{Introduction}
\label{sec:introduction}
Large language models (LLMs) deployed as conversational agents are constrained by fixed-size context windows.
As conversations grow, older turns must either be truncated, summarised, or selectively retained.
The dominant approach is \emph{generative compression}: an LLM periodically rewrites recent history into a shorter summary that replaces the original turns in the context.
While effective in practice, this strategy suffers from several scientific and engineering drawbacks.

\paragraph{Non-determinism.}
Two calls to the same LLM with the same history may produce different summaries, making the memory state non-reproducible and the system difficult to debug or audit.

\paragraph{Token cost.}
Every summarisation call consumes input tokens proportional to the history length and produces output tokens that re-enter the context.
In long-horizon deployments this cost can dwarf the cost of the agent task itself.

\paragraph{Semantic drift.}
When a summary is written, its meaning is fixed at write time.
If the NLP model or the conversational context evolves, the summary remains anchored to a stale interpretation.

\paragraph{Opacity.}
It is impossible to inspect why a particular fact was retained or discarded, because the decision is buried inside a neural generation step.

These limitations motivate a fundamentally different approach: replacing LLM-based memory management with a deterministic, mathematically defined pipeline.
DMF (Deterministic Memory Framework) is our answer to this challenge.

DMF is a CPU-first framework that manages conversational memory without invoking any LLM during the memory-management loop.
The system extracts numerical content signals and structured conversational cues from each interaction using classical NLP tools (spaCy ~\cite{honnibal2020spacy} for morpho-syntactic analysis and VADER ~\cite{hutto2014vader} for sentiment), computes a scalar Survival Score $\Omega$, applies an exponential decay modulated by the score itself, and makes pruning decisions based on deterministic rules over $\Omega_{\mathrm{eff}}$.
Evicted entries are archived as canonical raw records and may be accompanied by source-linked deterministic card projections; retrieval is performed at query time, so the recall semantics evolve with the current deterministic pipeline.

The key contributions of DMF are:

\begin{enumerate}
\item \textbf{Interaction-count decay.}
  We argue that for conversational memory, the relevant temporal axis is the number of newer interactions $\Delta n$, not wall-clock time.
  This choice makes the memory state a deterministic function of the message sequence alone.

\item \textbf{Score-dependent inertia.}
  The effective decay rate is modulated by the original survival score, so semantically rich entries resist decay more than marginal ones.

\item \textbf{Recall-time NLP.}
  Long-term memory preserves raw text records as the authoritative source; auxiliary cards remain deterministic projections linked back to those records. Semantic interpretation, filtering, and evidence assembly are performed at query time, decoupling storage stability from model evolution.

\item \textbf{Zero LLM calls in the memory loop.}
  Scoring, pruning, archival, retrieval, reranking, and evidence assembly are deterministic and require no generative model call.
\end{enumerate}

The remainder of the paper is organised as follows.
Section~\ref{sec:related} surveys related work.
Section~\ref{sec:architecture} describes the system architecture.
Section~\ref{sec:nlp} presents the NLP feature extraction pipeline.
Section~\ref{sec:scoring} derives the Survival Score.
Section~\ref{sec:decay} develops the temporal decay model.
Section~\ref{sec:pruning} details the pruning mechanisms.
Section~\ref{sec:social} discusses the social floor heuristic.
Section~\ref{sec:memory} describes the full memory lifecycle including LTM archival and recall.
Section~\ref{sec:implementation} covers implementation details.
Section~\ref{sec:benchmarks} presents benchmark results.
Sections~\ref{sec:conclusion} and~\ref{sec:future} give conclusions and future work.

\section{Related Work}
\label{sec:related}

\subsection{LLM-Based Memory Compression}

MemGPT ~\cite{packer2023memgpt} introduced the analogy between LLM context management and operating system memory hierarchies, using function calls to move information between a main context and external storage.
Memory transitions are managed by the LLM itself, which reads and writes a structured memory store.
DMF differs fundamentally: memory transitions are governed by a deterministic scoring function, not by LLM-generated instructions.

Mem0 ~\cite{chhikara2025mem0} proposes a production-grade memory layer in which an LLM extracts structured facts from conversations and stores them as semantic memories with conflict resolution.
The Mem0 evaluation on the \textit{LoCoMo} benchmark ~\cite{maharana2024locomo} demonstrates that targeted semantic storage outperforms full-context RAG in both accuracy and token efficiency.
DMF shares the goal of token efficiency but avoids LLM calls entirely during memory management, replacing fact extraction with deterministic signal analysis.

A-MEM ~\cite{li2025amem} builds an agentic memory system using Zettelkasten-inspired linking between memory notes, where an LLM assigns metadata and interconnects entries.
ReadAgent ~\cite{lee2024readagent} compresses conversations into short gists via LLM calls and uses them for retrieval.
MemoryBank ~\cite{zhong2023memorybank} combines episodic memory with a memory update mechanism inspired by the Ebbinghaus forgetting curve ~\cite{ebbinghaus1885memory}.
DMF takes direct inspiration from the exponential forgetting model but implements it deterministically rather than through neural updates.

\subsection{Retrieval-Augmented Generation}

Retrieval-Augmented Generation (RAG) ~\cite{lewis2020rag} augments LLM responses with documents retrieved from an external corpus.
Applied to conversational memory, RAG stores raw conversation turns and retrieves the most similar ones at query time ~\cite{gao2023retrieval}.
DMF extends this approach with a structured recall pipeline: query understanding, raw and card-based candidate channels, deterministic answerability reranking, topic-aware suppression, and source-grounded evidence assembly are applied before context rendering.

\subsection{Long-Horizon Conversation Benchmarks}

The LoCoMo dataset provides multi-session conversational data with structured question types designed to evaluate long-term memory recall.
LangMem ~\cite{langmem2024} and similar frameworks provide flexible memory APIs over LangChain ~\cite{langchain2023} but rely on LLM calls for memory distillation.
Our benchmark suite (Section~\ref{sec:benchmarks}) follows the structural design principles of LoCoMo while targeting specifically the deterministic properties of DMF: signal retention under noise, preference-update propagation, and correction-chain propagation.

\subsection{Cognitive Memory Models}

The Ebbinghaus forgetting curve ~\cite{ebbinghaus1885memory} characterises human memory retention as an exponential decay with time: $R(t) = e^{-t/S}$, where $S$ is the \emph{stability} of the memory trace.
The Spacing Effect and the ACT-R cognitive architecture ~\cite{anderson1997act} extend this model with activation-based retrieval thresholds.
DMF translates these ideas into a discrete, interaction-count domain: decay is exponential in $\Delta n$ (number of newer turns), and the stability analogue is the Survival Score $\Omega$ itself, which modulates the effective decay rate through the inertia term.

\section{System Architecture}
\label{sec:architecture}

DMF is organised into eight functional layers:

\begin{enumerate}
\item \textbf{Analysis Pipeline.} The \texttt{InteractionPipeline} coordinates the NLP engine, embedding engine, and interaction matrix. It produces the numerical content signals $(ID, |S|, E, D)$ together with the canonical embedding used by downstream memory components.
\item \textbf{Conversational Signal Layer.} A language-specific signal adapter extracts deterministic pragmatic cues: preference, constraint, correction, current-state and past-state markers, replacement patterns, query-like turns, acknowledgement-like turns, and narrow topic identity/value pairs.
\item \textbf{Scoring Engine.} The \texttt{ScoringEngine} computes the static Survival Score $\Omega$ from content signals, operational conversational signals, and structured provenance.
\item \textbf{TemporalMemory.} The active-memory manager stores \texttt{MemoryEntry} objects, applies interaction-count decay, resolves active visibility, enforces token-budget pruning, runs periodic cleanup, and orchestrates archival.
\item \textbf{LTM Backend.} Long-term memory stores canonical raw records in the vector backend. Vector-backed stores additionally expose semantic search over raw records and, when enabled, over auxiliary card projections.
\item \textbf{Card Projection Layer.} Evicted entries may be projected into deterministic \texttt{MemoryCard} objects representing conservative subject--predicate--object facts, preferences, constraints, relations, events, or current-state assertions. Cards are retrieval aids, not replacements for raw records.
\item \textbf{Structured Retrieval Stack.} Query understanding, multi-channel candidate generation, hard filtering, answerability-aware reranking, and evidence assembly convert the memory substrate into final \texttt{RetrievedEvidence}.
\item \textbf{Memory Facade.} The public API exposes \texttt{retrieve()} for structured evidence and \texttt{render\_context()} for prompt-ready context rendering.
\end{enumerate}

The central design invariant is \emph{source-canonical memory}: the authoritative archival object is always the raw interaction record.
Structured cards are deterministic auxiliary projections over this substrate; they improve retrieval through symbolic and semantic channels, but every card remains linked to its source record and is expanded back into raw supporting turns before prompt rendering.
This preserves the separation between archival and interpretation: stable source text remains recoverable, while derived semantics can be recomputed or revised by the current deterministic pipeline.

\begin{figure}[H]
\centering
\includegraphics[width=\textwidth]{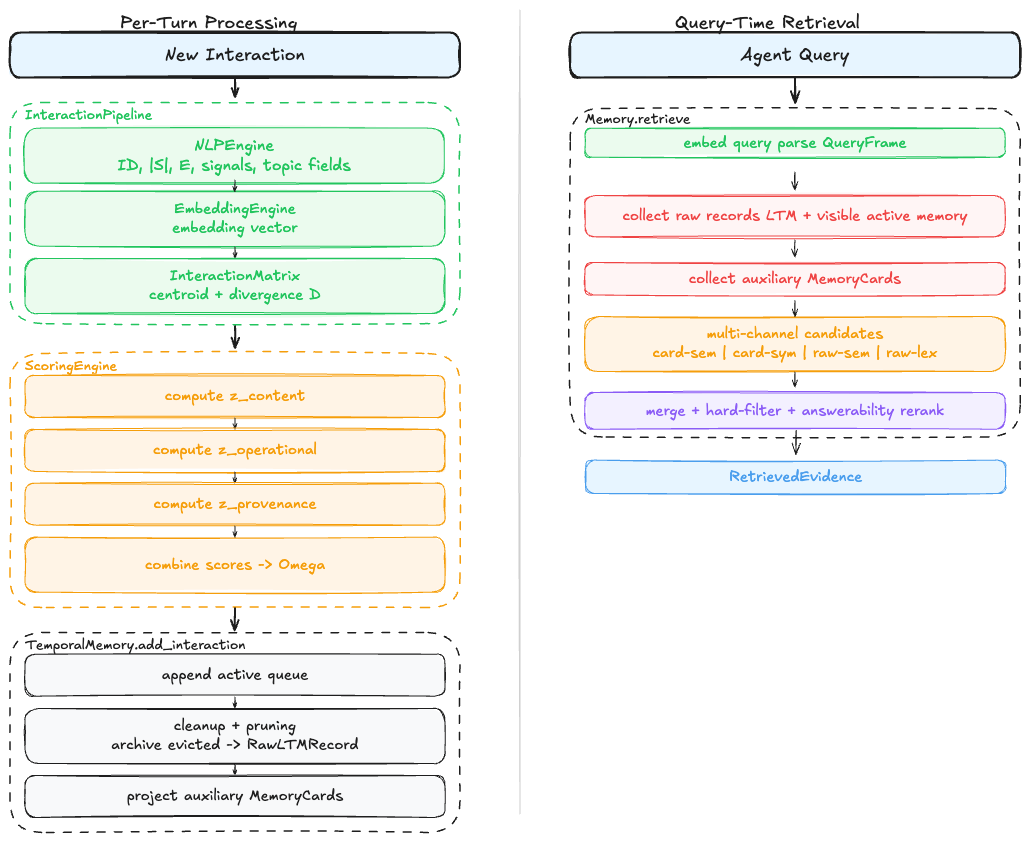}
\caption{DMF runtime pipeline.}
\label{fig:pipeline}
\end{figure}

In Figure \ref{fig:pipeline} we reported the DMF runtime pipeline. Per-turn processing (left) produces a Survival Score and updates active memory. Query-time retrieval (right) builds a deterministic evidence set from raw records, visible active memory, and optional structured card projections.

\section{NLP Feature Extraction}
\label{sec:nlp}

For each interaction text $t$, the NLP engine extracts three scalar content signals and a structured conversational-signal envelope, with no LLM involvement.
The scalar signals drive the content component of the Survival Score; the structured envelope is consumed by scoring, pruning, card projection, and retrieval.

\subsection{Information Density}

Information density $ID \in [0,1]$ measures the ratio of semantically load-bearing tokens to total tokens.
A token is considered semantic if its part-of-speech tag belongs to the set $\mathcal{P} = \{\texttt{NOUN},\, \texttt{VERB},\, \texttt{ADJ},\, \texttt{PROPN}\}$:

\begin{equation}
  ID(t) = \frac{|\{w \in t : \mathrm{pos}(w) \in \mathcal{P}\}|}{|t|}
  \label{eq:id}
\end{equation}

where $|t|$ denotes the total token count.
Purely phatic turns (e.g.\ ``OK'', ``Thanks!'') yield $ID \approx 0$ because interjections (\texttt{INTJ}) are excluded; technical or instructional turns yield $ID$ close to 1.

\subsection{Sentiment Magnitude}

Sentiment magnitude $|S| \in [0,1]$ is the absolute value of the VADER compound score ~\cite{hutto2014vader}:

\begin{equation}
  |S|(t) = |\mathrm{VADER\_compound}(t)|
  \label{eq:sentiment}
\end{equation}

The absolute value is used because both strongly positive and strongly negative interactions carry emotional salience relevant to memory retention.
VADER is chosen for its rule-based, deterministic character: the same text always produces the same score.

\subsection{Named Entity Count}

Named entity count $E \geq 0$ is the number of recognised named entities in $t$ using the spaCy standard NER model (categories: \texttt{PERSON}, \texttt{ORG}, \texttt{GPE}, \texttt{LOC}, \texttt{PRODUCT}, \texttt{EVENT}, and related types):

\begin{equation}
  E(t) = |\{e \in \mathrm{NER}(t)\}|
  \label{eq:entity}
\end{equation}

Named entities act as factual anchors; turns with many entities are more likely to introduce verifiable facts that should survive in memory.

\subsection{Entity Normalisation}

Because entity count is unbounded whereas information density and sentiment magnitude are already bounded, we apply a saturation normalisation with cap $E_{\mathrm{cap}}$:

\begin{equation}
  E_{\mathrm{norm}}(t) = \frac{\min(E(t),\; E_{\mathrm{cap}})}{E_{\mathrm{cap}}}
  \label{eq:enorm}
\end{equation}

The default $E_{\mathrm{cap}} = 5$ reflects the empirical observation that most conversational turns contain at most three to four entities, so saturation at five covers virtually all highly entity-rich turns without distorting the scale.

\subsection{Conversational and Pragmatic Signals}

In addition to the scalar content signals, DMF extracts a deterministic set of conversational signals:

\begin{equation}
  \mathcal{G}(t) =
  \{\mathrm{constraint},\;
    \mathrm{preference},\;
    \mathrm{current\_state},\;
    \mathrm{past\_state},\;
    \mathrm{correction},\;
    \mathrm{replacement},\;
    \mathrm{query\_like},\;
    \mathrm{ack\_like}\}.
  \label{eq:signals}
\end{equation}

These signals are produced by a language-specific rule adapter over the spaCy parse.
For English, the adapter combines phrase matching (e.g.\ ``I prefer'', ``currently'', ``actually'', ``do not''), narrow syntactic rules for first-person factual statements, quantitative markers, and simple replacement patterns such as ``not $x$ but $y$''.
The output includes auditable cue evidence and, when a stable normalisation is possible, a pair $(\mathrm{topic\_identity}, \mathrm{topic\_value})$.

The topic fields are intentionally conservative.
They are not a general-purpose semantic parser; they only capture patterns that can be normalised deterministically, such as preferences, constraints, and selected relation-like statements.
This conservative policy is important because topic fields are later used by pruning and retrieval to suppress superseded facts.

\subsection{Semantic Divergence}

The embedding engine maintains a moving centroid $\vec{C}$ as a weighted average of the embeddings of the last $W$ interactions (configurable window size):

\begin{equation}
  \vec{C}_n = \frac{\sum_{k=n-W}^{n-1} \vec{v}_k}{\left|\sum_{k=n-W}^{n-1} \vec{v}_k\right|}
  \label{eq:centroid}
\end{equation}

Semantic divergence $D \in [0, 2]$ is the cosine distance between the current interaction embedding $\vec{v}_n$ and the centroid:

\begin{equation}
  D(t) = 1 - \frac{\vec{v}_n \cdot \vec{C}_n}{\|\vec{v}_n\|\,\|\vec{C}_n\|}
  \label{eq:divergence}
\end{equation}

High divergence indicates a topic shift; this is penalised in the survival score to prevent off-topic noise from surviving pruning.
Note that a legitimate topic shift accompanied by high information density may still yield a high survival score, because $D$ enters as a penalty with weight $\delta < 0$, not as a hard gate.
The upper bound $2$ follows from clipping cosine similarity to $[-1,1]$ before applying the distance transform.

\section{Survival Score}
\label{sec:scoring}

\subsection{Linear Aggregation}

The Survival Score is computed from three additive pre-sigmoid channels: a content channel, an operational channel, and a provenance channel.
The content channel preserves the original four-signal formulation:

\begin{equation}
  z_{\mathrm{content}} =
    \alpha \cdot ID
    + \beta \cdot |S|
    + \gamma \cdot E_{\mathrm{norm}}
    + \delta \cdot D
  \label{eq:linear}
\end{equation}

where $\delta < 0$ makes divergence a penalty.
The default content weights are summarised in Table~\ref{tab:weights}.

\begin{table}[H]
\centering
\caption{Default content scoring weights and their rationale.}
\label{tab:weights}
\begin{tabular}{@{}clcl@{}}
\toprule
Symbol & Signal & Default & Rationale \\
\midrule
$\alpha$ & Information density $ID$ & $3.0$ & Dominant signal: content-rich messages must survive. \\
$\beta$ & Sentiment magnitude $|S|$ & $0.2$ & Weak supplement: sentiment alone should not dominate retention. \\
$\gamma$ & Entity density $E_{\mathrm{norm}}$ & $2.0$ & Strong supplement: entities are factual anchors. \\
$\delta$ & Divergence $D$ & $-2.5$ & Strong negative penalty: off-topic drift should be suppressed. \\
$x_0$ & Sigmoid midpoint & $1.5$ & Logistic midpoint for the combined pre-activation. \\
$E_{\mathrm{cap}}$ & Entity cap & $5$ & Most turns have 0--3 entities; 5 saturates the highly factual range. \\
\bottomrule
\end{tabular}
\end{table}

The operational channel gives explicit conversational memory cues a bounded positive contribution:

\begin{equation}
\begin{split}
  z_{\mathrm{op}} = \lambda_{\mathrm{op}} \cdot (&
    \eta_{\mathrm{con}} \mathbf{1}_{\mathrm{constraint}}
    + \eta_{\mathrm{pref}} \mathbf{1}_{\mathrm{preference}}
    + \eta_{\mathrm{cs}} \mathbf{1}_{\mathrm{current\_state}} \\
    &+ \eta_{\mathrm{corr}} \mathbf{1}_{\mathrm{correction}}
    + \eta_{\mathrm{repl}} \mathbf{1}_{\mathrm{replacement}}
    + \eta_{\mathrm{past}} \mathbf{1}_{\mathrm{past\_state}})
\end{split}
  \label{eq:operational}
\end{equation}

The default operational weights are $\lambda_{\mathrm{op}}=0.75$, $\eta_{\mathrm{con}}=1.20$, $\eta_{\mathrm{pref}}=0.70$, $\eta_{\mathrm{cs}}=0.60$, $\eta_{\mathrm{corr}}=0.90$, $\eta_{\mathrm{repl}}=0.50$, and $\eta_{\mathrm{past}}=0.0$.
This makes explicit constraints and corrections salient without allowing them to bypass decay or pruning.

The provenance channel accounts for structured metadata supplied by the caller:

\begin{equation}
  z_{\mathrm{prov}} =
    \kappa_{\mathrm{uc}}\mathbf{1}_{\mathrm{user\_correction}}
    + \kappa_{\mathrm{pu}}\mathbf{1}_{\mathrm{preference\_update}}
    + \kappa_{\mathrm{con}}\mathbf{1}_{\mathrm{constraint\_source}}
    - \kappa_{\mathrm{cb}}\mathbf{1}_{\mathrm{corrected\_by\_user}}.
  \label{eq:provenance}
\end{equation}

Unlike topic-supersession rules, provenance terms are local to the interaction being scored.
They let an adapter mark a turn as a user correction, a preference update, or a constraint without modifying the raw text.
The runtime default provenance weights are reported with the retrieval constants in Table~\ref{tab:runtime-defaults}.

\subsection{Logistic Projection}

The total pre-activation is

\begin{equation}
  z_{\mathrm{total}} = z_{\mathrm{content}} + z_{\mathrm{op}} + z_{\mathrm{prov}}.
  \label{eq:ztotal}
\end{equation}

The Survival Score is the logistic sigmoid of $z_{\mathrm{total}}$ centred at $x_0$:

\begin{equation}
  \Omega = \sigma(z_{\mathrm{total}} - x_0)
  = \frac{1}{1 + e^{-(z_{\mathrm{total}} - x_0)}}
  \label{eq:omega}
\end{equation}

The sigmoid is chosen because: (i) it maps the unbounded linear score to $(0,1)$; (ii) it is maximally sensitive near $x_0$, where most average turns cluster, so small signal differences create meaningful score differences; (iii) it saturates gracefully at both extremes rather than clipping.

\subsection{Calibration}

A representative average conversational turn has $ID \approx 0.40$, $|S| \approx 0.15$, $E_{\mathrm{norm}} \approx 0.20$, $D \approx 0.15$.
With no operational or provenance signal, the corresponding pre-activation is:

\begin{equation}
  z_{\mathrm{avg}} =
    z_{\mathrm{content}} =
    3.0(0.40) + 0.2(0.15) + 2.0(0.20) - 2.5(0.15) = 1.255
  \label{eq:zavg}
\end{equation}

With $x_0 = 1.5$:

\begin{equation}
  \Omega_{\mathrm{avg}} = \sigma(1.255 - 1.50) = \sigma(-0.245) \approx 0.439
  \label{eq:omegaavg}
\end{equation}

The average turn is slightly below the sigmoid midpoint, which is the desired operating point for the runtime configuration: neutral turns do not survive merely because they are emotionally marked, while operational and provenance cues can still lift durable conversational facts into the high-retention region.

\subsection{Representative Score Profiles}

Table~\ref{tab:profiles} shows representative turn types and their predicted scores.
For rows without an explicit operational signal, $z_{\mathrm{op}}=z_{\mathrm{prov}}=0$.

\begin{table}[H]
\centering
\caption{Representative survival score profiles.}
\label{tab:profiles}
\begin{tabular}{@{}lcccccc@{}}
\toprule
Profile & $ID$ & $|S|$ & $E_{\mathrm{norm}}$ & $D$ & $z_{\mathrm{total}}$ & $\Omega$ \\
\midrule
Technical fact (``Function must run in $O(n)$\ldots'') & 0.60 & 0.05 & 0.80 & 0.05 & 3.29 & 0.86 \\
User constraint (``Do not use external APIs'') & 0.35 & 0.05 & 0.20 & 0.05 & 2.24 & 0.68 \\
Content-rich, on-topic & 0.65 & 0.10 & 0.20 & 0.05 & 2.25 & 0.68 \\
Average turn & 0.40 & 0.15 & 0.20 & 0.15 & 1.26 & 0.44 \\
Emotional, off-topic drift & 0.15 & 0.85 & 0.10 & 0.70 & -0.93 & 0.08 \\
Empty / non-social & 0.00 & 0.00 & 0.00 & 0.00 & 0.00 & 0.18 \\
\bottomrule
\end{tabular}
\end{table}

The ordering reflects the runtime design intent: dense factual turns and explicit constraints survive, neutral average turns remain borderline, and emotionally marked but off-topic drift is strongly suppressed.

\section{Temporal Decay}
\label{sec:decay}

\subsection{Decay Axis: Interaction Count vs.\ Wall-Clock Time}

A natural first choice for the decay axis is wall-clock time $\Delta t_{\mathrm{wall}}$.
DMF explicitly rejects this choice. If the user pauses a conversation for an hour and then resumes, wall-clock decay would have silently erased memories even though no new information arrived.
This violates the principle that memory state should be a deterministic function of the conversational sequence.

For an interaction with identifier $i$ evaluated when the latest interaction identifier is $i_{\max}$:

\begin{equation}
  \Delta n = i_{\max} - i
  \label{eq:deltan}
\end{equation}

The newest interaction always has $\Delta n = 0$; older interactions have larger $\Delta n$.
The same sequence of messages always produces the same memory state, regardless of when it was played back.

\subsection{Exponential Decay Family}

Among the candidate decay families (linear, step, exponential), DMF adopts exponential decay for three reasons: it preserves a smooth ordering gradient at all values of $\Delta n$; it supports deterministic calibration via a single rate parameter; and it is composable—two decay processes applied in sequence remain exponential.

Without inertia, the decayed survival score would be:

\begin{equation}
  \Omega_{\mathrm{eff}}^{(0)}(\Delta n) = \Omega \cdot e^{-\lambda \Delta n}
  \label{eq:plain_decay}
\end{equation}

where $\lambda > 0$ is the base decay rate.

\subsection{Score-Dependent Inertia}

Uniform decay ($\lambda$ identical for all entries) is mathematically sound but operationally undesirable: a highly salient anchor and a marginal filler message would lose the same fraction per turn, eroding the relative advantage of high-quality entries.

DMF introduces a \emph{score-dependent inertia} multiplier:

\begin{equation}
  \mu(\Omega) = 1 - \eta \cdot \Omega
  \label{eq:inertia}
\end{equation}

where $\eta \in [0,1)$ is the inertia strength.
As $\Omega$ increases, $\mu$ decreases, so high-quality entries decay at a lower effective rate.

\subsection{Full Decay Law}

Combining equations~\eqref{eq:plain_decay} and~\eqref{eq:inertia}:

\begin{equation}
  \boxed{
    \Omega_{\mathrm{eff}}(\Delta n) = \Omega \cdot \exp\!\bigl(-\lambda \cdot (1 - \eta \Omega) \cdot \Delta n\bigr)
  }
  \label{eq:decay}
\end{equation}

\textbf{Properties.}
\begin{itemize}
\item $\Omega_{\mathrm{eff}}(0) = \Omega$: the score is unchanged when the interaction is brand new.
\item $\Omega_{\mathrm{eff}}$ is strictly monotonically decreasing in $\Delta n$ for $\Omega > 0$.
\item $\Omega_{\mathrm{eff}} \in (0, \Omega]$ for all $\Delta n \geq 0$.
\item The effective decay rate $\lambda_{\mathrm{eff}} = \lambda(1-\eta\Omega)$ is a smooth function of $\Omega$, avoiding the tier-boundary discontinuities that a step-based inertia rule would introduce.
\end{itemize}

\subsection{Effective Half-Life}

For a fixed starting score $\Omega$, the number of turns until $\Omega_{\mathrm{eff}}$ is halved is:

\begin{equation}
  t_{1/2}(\Omega) = \frac{\ln 2}{\lambda \cdot (1 - \eta\Omega)}
  \label{eq:halflife}
\end{equation}

Table~\ref{tab:halflife} gives numerical values for the default parameters $\lambda = 0.035$, $\eta = 0.5$.

\begin{table}[H]
\centering
\caption{Effective half-lives for representative starting scores ($\lambda=0.035$, $\eta=0.5$).}
\label{tab:halflife}
\begin{tabular}{@{}rrrrl@{}}
\toprule
$\Omega$ & $\mu = 1 - 0.5\Omega$ & $\lambda_{\mathrm{eff}}$ & Half-life (turns) & Tier at creation \\
\midrule
0.87 & 0.565 & 0.0198 & 35.0 & \textsc{healthy} \\
0.72 & 0.640 & 0.0224 & 30.9 & \textsc{unstable} \\
0.50 & 0.750 & 0.0263 & 26.4 & \textsc{unstable} \\
0.35 & 0.825 & 0.0289 & 24.0 & \textsc{unstable} \\
0.25 & 0.875 & 0.0306 & 22.6 & \textsc{critical} \\
0.18 & 0.910 & 0.0319 & 21.8 & \textsc{critical} \\
\bottomrule
\end{tabular}
\end{table}

The gap in half-life between the highest- and lowest-scoring entries ($\approx 35$ vs.\ $\approx 22$ turns) is intentional but bounded: high-quality entries persist substantially longer, but no entry is immortal.

\subsection{Memory Tier Classification}

The effective score $\Omega_{\mathrm{eff}}$ is used to classify each active entry into one of three tiers:

\begin{equation}
  \mathrm{tier}(\Omega_{\mathrm{eff}}) =
  \begin{cases}
    \textsc{healthy}  & \text{if } \Omega_{\mathrm{eff}} > \tau_H \\
    \textsc{unstable} & \text{if } \tau_C < \Omega_{\mathrm{eff}} \leq \tau_H \\
    \textsc{critical} & \text{if } \Omega_{\mathrm{eff}} \leq \tau_C
  \end{cases}
  \label{eq:tiers}
\end{equation}

Default thresholds: $\tau_H = 0.75$ (\textsc{healthy} floor) and $\tau_C = 0.3$ (\textsc{critical} ceiling).
These thresholds serve as classification labels for logging and pruning eligibility checks; they do not alter the decay equation itself.

\section{Pruning Mechanisms}
\label{sec:pruning}

DMF operates two distinct removal regimes: \emph{periodic hard-kill} and \emph{budget-pressure pruning}.
They are semantically distinct and not interchangeable.

\subsection{Periodic Hard-Kill}

At configurable intervals, a cleanup sweep removes every active entry whose effective score has fallen below the hard-kill floor $\Omega_{\mathrm{kill}}$:

\begin{equation}
  \text{evict}(i) \iff \Omega_{\mathrm{eff},i} < \Omega_{\mathrm{kill}}
  \label{eq:hardkill}
\end{equation}

Default: $\Omega_{\mathrm{kill}} = 0.05$.
This rule applies to all entries regardless of tier; even a formerly \textsc{healthy} entry is removed if it has decayed below the floor.
There is no promotion step and no summarisation: the evicted entry is archived as a raw LTM record.

\subsection{Budget-Pressure Pruning}

When the active queue's token count exceeds the configured budget $\tau_{\mathrm{tok}}$ (default: 4096 tokens), DMF must remove entries to restore compliance.

\subsubsection{Candidate Set}

Only entries with $\Omega_{\mathrm{eff}} \leq \tau_H$ are eligible:

\begin{equation}
  \mathcal{C} = \{i : \Omega_{\mathrm{eff},i} \leq \tau_H\}
  \label{eq:candidates}
\end{equation}

\textsc{healthy} entries ($\Omega_{\mathrm{eff}} > \tau_H$) are protected from budget-pressure pruning.
This is a design policy, not a consequence of the decay equation.

\subsubsection{Retention Bonus}

For each entry in $\mathcal{C}$, a retention bonus $B_{\mathrm{ret}}$ is computed from its conversational signals:

\begin{equation}
  B_{\mathrm{ret}} = \rho_{\mathrm{con}} \cdot \mathbf{1}_{\mathrm{constraint}}
                   + \rho_{\mathrm{pref}} \cdot \mathbf{1}_{\mathrm{preference}}
                   + \rho_{\mathrm{cs}} \cdot \mathbf{1}_{\mathrm{current\_state}}
                   + \rho_{\mathrm{corr}} \cdot \mathbf{1}_{\mathrm{correction}}
                   + \rho_{\mathrm{repl}} \cdot \mathbf{1}_{\mathrm{replacement}}
  \label{eq:retention}
\end{equation}

Default bonus values: $\rho_{\mathrm{con}} = 0.20$, $\rho_{\mathrm{pref}} = 0.10$, $\rho_{\mathrm{cs}} = 0.10$, $\rho_{\mathrm{corr}} = 0.15$, $\rho_{\mathrm{repl}} = 0.08$.

These bonuses protect operationally significant turns (user constraints, preferences, corrections) from being evicted under token pressure, even if their information density is low.

\subsubsection{Topic-Supersession Penalty}

When a newer interaction is identified as superseding an older one on the same topic (e.g.\ a preference update), the older entry receives a penalty:

\begin{equation}
  P_{\mathrm{sup}} = p_{\mathrm{sup}} \cdot \mathbf{1}_{\mathrm{superseded}}
  \label{eq:penalty}
\end{equation}

Default: $p_{\mathrm{sup}} = 0.35$.
A superseded entry becomes easier to evict under pressure even if its raw $\Omega_{\mathrm{eff}}$ is not minimal.

\subsubsection{Pruning Score and Eviction Rule}

The effective pruning score for entry $i$ is:

\begin{equation}
  S_{\mathrm{prune},i} = \Omega_{\mathrm{eff},i} + B_{\mathrm{ret},i} - P_{\mathrm{sup},i}
  \label{eq:sprune}
\end{equation}

The eviction rule under budget pressure is:

\begin{equation}
  i^* = \operatorname*{argmin}_{i \in \mathcal{C}} S_{\mathrm{prune},i}
  \label{eq:evict}
\end{equation}

with interaction ID used as a deterministic tie-breaker.
The entry $i^*$ is removed from the active queue and archived to LTM; the process repeats until the token budget is satisfied.

\textbf{Key distinction.}
Budget pruning is not equivalent to sorting by $\Omega_{\mathrm{eff}}$ alone.
An entry with a low $\Omega_{\mathrm{eff}}$ but a strong retention bonus ($+0.20$ for constraints) may rank higher than an entry with a slightly higher $\Omega_{\mathrm{eff}}$ but no retention bonus.
Conversely, a superseded entry receives a penalty that accelerates its eviction regardless of its current score.

\section{Social Floor}
\label{sec:social}

\subsection{Problem: Phatic Turns}

In natural conversation, a significant fraction of turns serve a purely phatic function: greetings (``Hi!''), acknowledgments (``OK'', ``Got it''), gratitude (``Thanks!''), and pleasantries (``Awesome!'').
These turns carry little or no propositional content; under the base scoring formula of Equation~\eqref{eq:omega}, they typically receive $\Omega \approx 0.18$ (the all-zero-signal boundary case):

\begin{equation}
  \Omega_{\mathrm{phatic}} = \sigma(0 - x_0) = \sigma(-1.5) \approx 0.18
  \label{eq:phatic}
\end{equation}

Phatic turns would be the first candidates for pruning and would be evicted rapidly.
While this is arguably correct behaviour, premature eviction can create coherence gaps in the active context.

\subsection{Heuristic: Short-Message Keyword Gate}

DMF applies a \emph{Social Floor} heuristic to raise the minimum score of recognised phatic turns:

\begin{equation}
  \mathrm{is\_social}(t) = \bigl[\mathrm{word\_count}(t) \leq 6\bigr]
                           \;\wedge\;
                           \bigl[\exists\, w \in t : w \in \mathcal{K}_{\mathrm{social}}\bigr]
  \label{eq:social}
\end{equation}

where $\mathcal{K}_{\mathrm{social}}$ is a curated keyword list (``thanks'', ``ok'', ``hello'', ``hi'', ``great'', etc.).

The length cap of six words prevents false positives: a sentence such as ``I really appreciate your comprehensive analysis of the algorithm'' contains a gratitude keyword but is clearly a substantive, information-dense turn that should be scored normally.

The social floor is applied only when the raw $\Omega$ is already below an activation threshold $\tau_{\mathrm{soc}}$ (default: $0.40$), and raises it only to a lower minimum floor $\Omega_{\mathrm{soc}}^{\min}$ (default: $0.25$):

\begin{equation}
  \Omega_{\mathrm{final}} =
  \begin{cases}
    \max(\Omega,\; \Omega_{\mathrm{soc}}^{\min}) & \text{if } \mathrm{is\_social}(t) \wedge \Omega < \tau_{\mathrm{soc}} \\
    \Omega & \text{otherwise}
  \end{cases}
  \label{eq:socialfloor}
\end{equation}

This ensures the social floor has no effect on turns that naturally score above $\tau_{\mathrm{soc}}$.

\subsection{Why Not Sentiment-Based?}

VADER returns near-zero sentiment for many common acknowledgments (``ok'', ``sure'', ``noted'').
Keyword matching is more reliable for these cases.
Sentiment-based gating would miss precisely the turns that need the floor most.

\section{Memory Lifecycle}
\label{sec:memory}

\subsection{Active Memory}

\texttt{TemporalMemory} maintains a bounded active queue.
Each \texttt{MemoryEntry} stores source text, an \texttt{AnalysisReport} (containing $ID$, $|S|$, $E$, $D$, $\Omega$, status, conversational signals, and topic fields), embedding vector, token count, timestamp, provenance, and lineage.

At each new interaction, the pipeline:
\begin{enumerate}
\item Extracts scalar content signals $(ID, |S|, E)$ and conversational signals via the NLP engine.
\item Computes the embedding $\vec{v}$ and divergence $D$ via the embedding engine.
\item Computes $\Omega$ via the scoring engine, including content, operational, and provenance channels.
\item Appends the entry to the active queue.
\item At periodic intervals, runs hard-kill cleanup.
\item Applies the token budget check; if exceeded, runs budget-pressure pruning.
\end{enumerate}

\subsection{Long-Term Memory Archival}

When an entry is evicted from active memory—either by hard-kill or budget-pressure pruning—it is archived as a \texttt{RawLTMRecord}:

\begin{equation}
  \texttt{RawLTMRecord} = \{
    \mathtt{record\_id},\;
    \mathtt{interaction\_id},\;
    \mathtt{role},\;
    \mathtt{text},\;
    \mathtt{created\_at},\;
    \mathtt{provenance}
  \}
  \label{eq:rawrecord}
\end{equation}

Crucially, the \texttt{RawLTMRecord} excludes scores, topic metadata, signals, embeddings, and prompt renderings.
The rationale is that all of these derived fields must be recomputed at recall time to follow the current NLP engine.
Storing only the immutable raw text and provenance ensures that the LTM remains stable across engine upgrades.

In the vector-backed LTM backend, the \texttt{RawLTMRecord} is indexed for semantic search.

DMF also projects evicted entries into auxiliary \texttt{MemoryCard} objects:

\begin{equation}
\begin{aligned}
  \texttt{MemoryCard} = \{&
    \mathtt{card\_id},\;
    \mathtt{kind},\;
    \mathtt{subject},\;
    \mathtt{predicate},\;
    \mathtt{object},\\
    &\mathtt{qualifiers},\;
    \mathtt{time\_anchor},\;
    \mathtt{validity},\;
    \mathtt{provenance}
  \}.
\end{aligned}
  \label{eq:memorycard}
\end{equation}

Cards are conservative deterministic projections of raw entries.
They may encode preferences, constraints, corrections, current-state facts, relations, or simple events, but they remain auxiliary: every card carries a \texttt{source\_record\_id} and must be expanded back into raw supporting evidence before prompt rendering.

\subsection{Recall Pipeline}

When an agent queries the memory system through the public facade, the structured recall pipeline executes the following steps:

\paragraph{Step 1: Query understanding.}
The query text $q$ is embedded to $\vec{v}_q$ and parsed into a \texttt{QueryFrame}.
The frame contains deterministic query metadata: entities, aliases, subject focus, predicate focus, answer type, temporal intent, currentness focus, polarity, and filters.

\begin{equation}
  Q = \operatorname{parse}(q, \vec{v}_q)
  \label{eq:queryframe}
\end{equation}

\paragraph{Step 2: Evidence substrate construction.}
The retriever reads archived raw records from LTM and visible active records from \texttt{TemporalMemory}.
When a card store is available, it also reads auxiliary \texttt{MemoryCard} projections.
The active-memory guard suppresses records invalidated by lineage, user corrections, or newer topic winners before they can be rendered as context.

\paragraph{Step 3: Multi-channel candidate generation.}
Candidate generation combines four deterministic channels:
\begin{itemize}
\item \textbf{Raw semantic retrieval}: vector search over raw LTM records.
\item \textbf{Raw lexical retrieval}: normalised token and lemma overlap over raw records.
\item \textbf{Card semantic retrieval}: vector search over card projections when a card index exists.
\item \textbf{Card symbolic retrieval}: lookup by query entities, subject focus, predicate focus, answer type, and card fields.
\end{itemize}

The candidate pool is the deterministic merge of these channels:

\begin{equation}
  \mathcal{P}(Q) =
    \operatorname{dedupe}\bigl(
      \mathcal{C}_{\mathrm{raw\_sem}}
      \cup \mathcal{C}_{\mathrm{raw\_lex}}
      \cup \mathcal{C}_{\mathrm{card\_sem}}
      \cup \mathcal{C}_{\mathrm{card\_sym}}
    \bigr).
  \label{eq:candidatepool}
\end{equation}

Preliminary hard filters remove duplicate active records, invalidated or superseded records, evidence-type mismatches, and hard entity mismatches.

\paragraph{Step 4: Answerability reranking.}
Surviving candidates are reranked by deterministic answerability features rather than by vector similarity alone:

\begin{equation}
  A(e, Q) =
    \sum_m w_m f_m(e,Q)
    -
    \sum_\ell \pi_\ell p_\ell(e,Q).
  \label{eq:answerability}
\end{equation}

The positive features include semantic similarity, entity overlap, subject match, predicate match, query-specificity match, temporal compatibility, current-state compatibility, active/not-superseded status, answer-span likelihood, and evidence-chain availability.
The penalties suppress stale facts, acknowledgement-like social turns, and candidates that match only a topic name without answer-bearing content.

\paragraph{Step 5: Evidence assembly.}
The highest-ranked candidates are truncated to the configured final recall limit.
Winning cards are expanded into auditable raw support turns: the source turn, historical or supersession links when relevant, and optionally neighbouring turns.
The final output is a list of \texttt{RetrievedEvidence} objects whose provenance records the contributing retrieval channels and support record identifiers.

\paragraph{Step 6: Context rendering.}
For prompt injection, retrieved evidence is rendered before the visible active conversation:

\begin{lstlisting}[language={}]
=== LONG-TERM MEMORY (RECALLED) ===
=== STRUCTURED EVIDENCE ===
<answerability-ranked evidence items>

=== RAW SUPPORTING EVIDENCE ===
<source-linked raw turns>

=== ACTIVE CONVERSATION ===
<visible active queue entries>
\end{lstlisting}

The recalled section is omitted when no candidate survives the retrieval and reranking pipeline.

\subsection{Memory Lineage}

Active entries carry a lightweight \texttt{MemoryLineage} relation envelope that records structured relations to other entries.
Supported relation types: \texttt{supersedes}, \texttt{conflicts\_with}, \texttt{corrects}, \texttt{invalidates}.
Lineage information is used during pruning (supersession penalty), active-context suppression, recall-time suppression, and card-support expansion.
It is not stored inside \texttt{RawLTMRecord}; source-level archival remains minimal.

\section{Implementation}
\label{sec:implementation}

\subsection{Technology Stack}

DMF has been implemented in Python and designed for CPU-first deployment. The framework’s source code is available on \href{https://github.com/matstech/dmf}{github.com/matstech/dmf}.

\subsection{Configuration Surface}

All tunable parameters are exposed in \texttt{config/dmf\_settings.toml}.
The main sections are:
\begin{itemize}
\item \texttt{[scoring\_weights]}: content weights $\alpha, \beta, \gamma, \delta$, sigmoid midpoint $x_0$, entity cap $E_{\mathrm{cap}}$, social-floor parameters, operational weights, and provenance boosts.
\item \texttt{[temporal\_decay]}: $\lambda$, $\eta$, $\Omega_{\mathrm{kill}}$.
\item \texttt{[memory\_tiers]}: $\tau_C$, $\tau_H$.
\item \texttt{[capacity]}: token budget, pruning frequency, centroid window.
\item \texttt{[pruning\_priority]}: $\rho_{\mathrm{con}}, \rho_{\mathrm{pref}}, \rho_{\mathrm{cs}}, \rho_{\mathrm{corr}}, \rho_{\mathrm{repl}}, p_{\mathrm{sup}}$.
\item \texttt{[ltm]}: backend type, storage parameters, recall limit, distance threshold, and optional card-index settings.
\item \texttt{[retrieval]}: candidate prefetch sizes, channel gates, final recall limit, and evidence-expansion controls.
\end{itemize}

Table~\ref{tab:runtime-defaults} reports the additional runtime constants that affect provenance scoring and structured retrieval.

\begin{table}[H]
\centering
\small
\caption{Additional runtime defaults.}
\label{tab:runtime-defaults}
\begin{tabular}{@{}ll@{}}
\toprule
Component & Defaults \\
\midrule
Provenance boosts & $(0.15,\;0.10,\;0.10,\;0.0)$ for user correction, preference update, constraint source, corrected-by-user. \\
Recall limits & raw recall $5$; distance threshold $0.7$; raw/card/symbolic prefetch $16/32/16$; final recall $5$. \\
Evidence expansion & max support turns per card $3$; historical supersession on; neighbour expansion off. \\
Answerability weights & features $(1.0,1.0,1.2,1.15,1.2,0.8,0.9,1.0,1.0,0.4)$; penalties $(1.1,0.9,1.2)$. \\
Candidate scores & raw lexical $|Q \cap C|/|Q|$; symbolic card weights $(0.9,0.55,1.2,1.35,0.2)/4.2$. \\
\bottomrule
\end{tabular}
\end{table}

The following policies are intentionally not config-driven, as they form the fixed algorithmic core:
English cue lists and pragmatic heuristics; conservative card-projection rules; answerability feature definitions; topic-supersession and lineage-suppression rules.

\subsection{Public API}

The public memory facade exposes two methods:

\begin{lstlisting}[language=Python]
# Retrieve final structured evidence
evidence: list[RetrievedEvidence] = memory.retrieve(
    query_text="..."
)

# Render retrieved evidence as prompt-ready text
context: str = memory.render_context(
    query_text="..."
)
\end{lstlisting}

All query embedding, query parsing, candidate generation, answerability reranking, and evidence assembly are internal to the facade.
Runtime integrations that require the complete prompt surface prepend the rendered retrieved evidence to the active-conversation block produced by \texttt{TemporalMemory}.

\subsection{Computational Complexity}

\paragraph{Per-turn cost.}
NLP analysis: $O(L)$ where $L$ is the text length (spaCy linear pass).
Embedding: one forward pass of the embedding model; CPU inference is viable for models of dimension 384--768.
Scoring: $O(1)$.
Pruning: $O(|\mathcal{C}|)$ where $|\mathcal{C}| \leq$ active queue size.
Total LLM calls: \textbf{zero}.

\paragraph{Query-time cost.}
Query understanding: $O(L_q)$ plus one embedding forward pass for the query.
Raw semantic retrieval is $O(|\mathrm{LTM}| \cdot d)$ in the exact case and approximately $O(\log |\mathrm{LTM}|)$ with ANN indexing.
Raw lexical retrieval is $O(R \cdot L)$ over the readable raw-record set $R$ when enabled.
Card semantic retrieval follows the same vector-search bound over the card index; card symbolic lookup is $O(M)$ over available cards $M$ when implemented as a deterministic linear scan.
Candidate merging is $O(P \log P)$ for candidate pool size $P$; answerability reranking is $O(P)$; evidence expansion is $O(K + S)$ for final winners $K$ and support records $S$.

The critical result is that memory management consumes \emph{zero LLM tokens}.
Compared with summarisation-based memory, which repeatedly spends $O(H \cdot L)$ input tokens over history of length $H$, DMF uses deterministic CPU-bound processing and embedding/vector operations outside the generative model loop.

\section{Benchmarks}
\label{sec:benchmarks}

We created a benchmark pipeline execution that stores a session of questions of answers in DMF and then evaluate the response using LLM-as-judge against a goldenset. In Appendix \ref{sec:appendix} we reported the prompts used by the LLM-as-judge. We used the dataset of \textit{LoCoMo} and \textit{LongMemEval} ~\cite{wu2025longmemeval} for the goldenset. For \textit{LongMemEval} we used a  random sampling of 10 questions for all the 6 categories, we called this dataset \textit{LongMemEval-10}. We compared the execution of DMF with Mem0.

We used ChromaDB ~\cite{chromadb2023} as vector database, BAAI/bge-small-en-v1.5 ~\cite{xiao2023bge} as embedding model, gpt-4.1-mini \footnote{https://openai.com/index/gpt-4-1/} as LLM for the conversations, and GPT-5-mini ~\cite{singh2026openaigpt5card} as LLM-as-judge.
We executed the benchmark using an Intel(R) Xeon(R) Platinum 8168 CPU @ 2.70GHz with 16 GB of RAM, running on Linux 6.8.0-111-generic.

The code of the benchmark is available on \href{https://github.com/matstech/dmf-benchmarks}{https://github.com/matstech/dmf-benchmarks}. 

\subsection{LoCoMo}

We evaluated the accuracy, the execution time and the token consumption of DMF and Mem0.

We reported the \textit{LoCoMo} scores in Table \ref{tab:results-locomo}.

\begin{table}[ht]
\centering
\caption{LoCoMo scores}
\label{tab:results-locomo}
\begin{tabular}{llrrrr}
\hline
\textbf{Framework} & \textbf{Group} & \textbf{Count} & \textbf{Avg. Judge Score} & \textbf{Exact Match} & \textbf{Token F1} \\
\hline
\textbf{DMF}  & \textbf{overall} & \textbf{1540} & \textbf{0.7753} & \textbf{0.1039} & \textbf{0.3720} \\
DMF  & multi-hop & 282  & 0.7943 &  0.0355 & 0.2729 \\
DMF  & temporal & 321  & 0.7072 & 0.0218 & 0.2889 \\
DMF  & open-domain & 96   & 0.5729 & 0.0417 & 0.1598 \\
DMF  & single-hop & 841  & 0.8181 & 0.1653 & 0.4612 \\
\textbf{Mem0} & \textbf{overall} & \textbf{1540} & \textbf{0.6883} & \textbf{0.0747} & \textbf{0.3142} \\
Mem0 & multi-hop & 282  & 0.7943 & 0.0390 & 0.2951 \\
Mem0 & temporal & 321  & 0.1526 & 0.0093 & 0.0897 \\
Mem0 & open-domain & 96   & 0.5625 & 0.0521 & 0.1629 \\
Mem0 & single-hop & 841  & 0.8716 & 0.1141 & 0.4235 \\
\hline
\end{tabular}
\end{table}

The meanings of the columns are described as follows:

\paragraph{Group.}
The \textit{Group} are the LoCoMo groups divided in the following categories:

\begin{itemize}
\item \textbf{multi-hop}, the answer requires combining information from multiple sessions or dialogue parts.
\item \textbf{temporal}, the answer requires understanding dates, time order, duration, or temporal clues.
\item \textbf{open-domain}, the answer requires combining conversation information with external/common knowledge.
\item \textbf{single-hop}, the answer is directly stated in one session or one part of the conversation.
\end{itemize}

\paragraph{Count.}
\textit{Count} is the number of evaluation rows in the slice.

\paragraph{Avg. Judge Score.}
\textit{Avg. Judge Score} is the average of the correct answers evaluated by a LLM-judge (accuracy). The judge score is evaluated using a binary approach: the answer for the evaluation is always \textit{yes} or \textit{no}.

\paragraph{Exact Match.}
\textit{Exact Match} is the mean normalized exact-answer match.

\paragraph{Token F1.}
\textit{Token F1} is the mean token-overlap F1 against the gold answer.

\begin{figure}[ht]
\centering
\begin{tikzpicture}

\begin{groupplot}[
    group style={
        group size=2 by 2,
        horizontal sep=1.8cm,
        vertical sep=2.2cm
    },
    ybar,
    /pgf/bar width=6pt,
    width=0.48\textwidth,
    height=5.2cm,
    ymin=0,
    symbolic x coords={overall,multi-hop,temporal,open-domain,single-hop},
    xtick=data,
    legend style={
        at={(1.50,-0.60)},
        anchor=center,
        legend columns=2
    },
    tick label style={font=\small},
    label style={font=\small},
    title style={font=\small},
    xticklabel style={
        rotate=45,
        anchor=east
    }
]

\nextgroupplot[title={Avg. Judge Score}]
\addplot coordinates {
    (overall,0.7753)
    (multi-hop,0.7943)
    (temporal,0.7072)
    (open-domain,0.5729)
    (single-hop,0.8181)
};
\addplot coordinates {
    (overall,0.6883)
    (multi-hop,0.7943)
    (temporal,0.1526)
    (open-domain,0.5625)
    (single-hop,0.8716)
};
\legend{DMF, Mem0}

\nextgroupplot[title={Exact Match}]
\addplot coordinates {
    (overall,0.1039)
    (multi-hop,0.0355)
    (temporal,0.0218)
    (open-domain,0.0417)
    (single-hop,0.1653)
};
\addplot coordinates {
    (overall,0.0747)
    (multi-hop,0.0390)
    (temporal,0.0093)
    (open-domain,0.0521)
    (single-hop,0.1141)
};

\nextgroupplot[title={Token F1}]
\addplot coordinates {
    (overall,0.3720)
    (multi-hop,0.2729)
    (temporal,0.2889)
    (open-domain,0.1598)
    (single-hop,0.4612)
};
\addplot coordinates {
    (overall,0.3142)
    (multi-hop,0.2951)
    (temporal,0.0897)
    (open-domain,0.1629)
    (single-hop,0.4235)
};

\end{groupplot}

\end{tikzpicture}
\caption{DMF vs. Mem0 metrics using LoCoMo}
\label{fig:locomo-metrics}
\end{figure}
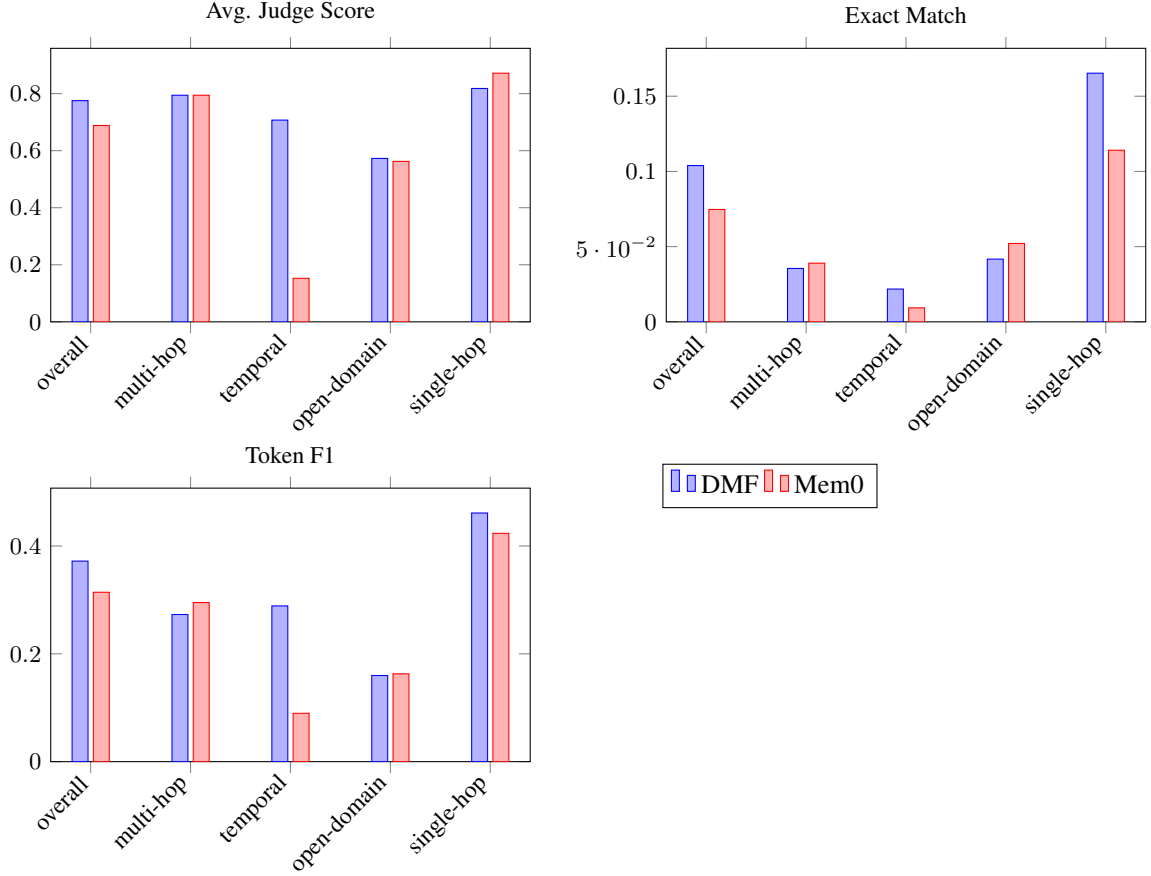

DMF wins over Mem0 for all the metrics: \textbf{+12\%} accuracy, \textbf{+39\%} exact match, \textbf{+18\%} Token F1 (Figure \ref{fig:locomo-metrics}).

On the \textbf{temporal reasoning} group of LoCoMo, DMF performs 4× better than Mem0 that is known to face challenges with temporal reasoning\footnote{https://mem0.ai/blog/the-token-efficient-memory-algorithm-now-has-temporal-reasoning}. DMF achieves better performance in this setting because it preserves absolute timestamps and conversational order as part of both the memory representation and the final prompt. In contrast, Mem0 tends to expose synthesized memories, in which relative dates may have already been incorrectly resolved. As a result, the final LLM receives temporally grounded evidence from DMF, whereas the memory provided by Mem0 may already contain temporal distortions. 

We evaluated the token consumption of DMF and Mem0 (Table \ref{tab:locomo-token}). Since DMF does not use any LLM to generate the memory context, the token consumption is 0 (Mem. Total).

\begin{table}[ht]
\centering
\caption{Framework token usage with LoCoMo}
\label{tab:locomo-token}
\resizebox{\textwidth}{!}{%
\begin{tabular}{lrrrrrr}
\hline
\textbf{Framework} &
\textbf{E2E Input} &
\textbf{Prompt} &
\textbf{Mem. Prompt} &
\textbf{Mem. Compl.} &
\textbf{Mem. Total} &
\textbf{Rec. E2E} \\
\hline
DMF & 10575032 & 10575032 & 0 & 0 & 0 & 10599346 \\
Mem0 & 56170922 & 6644319 & 49526603 & 456158 & 49982761 & 56651414 \\
\hline
\end{tabular}%
}
\end{table}

The meanings of the columns in Table \ref{tab:locomo-token} are described as follows:

\paragraph{E2E Input.}
The \textit{E2E Input} column reports the total number of input tokens used end-to-end. This includes the tokens sent to the final answerer as well as any internal memory prompt tokens used by the framework. It excludes completion tokens.

\paragraph{Prompt.}
The \textit{Ans. Prompt} column reports the total number of tokens sent to the answer-generation model. These are the tokens used in the final question-answering prompt, including the question and the retrieved or constructed context.

\paragraph{Mem. Prompt.}
The \textit{Mem. Prompt} column reports the total number of prompt tokens used internally by the memory framework before final answer generation. These tokens are used for memory retrieval, memory reasoning, summarization, or other framework-specific internal operations.

\paragraph{Mem. Compl.}
The \textit{Mem. Compl.} column reports the total number of completion tokens generated during internal memory operations. A value of 0 indicates that no internal memory completion token usage was recorded.

\paragraph{Mem. Total.}
The \textit{Mem. Total} column reports the total token usage of the internal memory component. It is the sum of internal memory prompt tokens and internal memory completion tokens.

\paragraph{Rec. E2E.}
The \textit{Rec. E2E} column reports the recorded end-to-end token total. This value is obtained by summing the recorded total end-to-end token values across evaluation rows.

DMF uses 5× fewer tokens than Mem0 (Rec. E2E). However, DMF uses a prompt that contains 1.5 times more tokens than Mem0. This is necessary because DMF does not perform any summarization. The size of the memory context is a configurable parameter, allowing users to balance token consumption against the desired level of accuracy.

\subsection{LongMemEval-10}

We reported the \textit{LongMemEval-10} scores in Table \ref{tab:results-longmemeval}.

\begin{table}[ht]
\centering
\caption{LongMemEval-10 scores}
\label{tab:results-longmemeval}
\begin{tabular}{llrrr}
\hline
\textbf{Framework} & \textbf{Group} & \textbf{Count} & \textbf{Judge Count} & \textbf{Avg. Judge Score} \\
\hline
\textbf{DMF}  & \textbf{overall} & \textbf{60} & \textbf{60} & \textbf{0.717} \\
DMF  & knowledge-update & 10  & 10 & 1.000 \\
DMF  & multi-session & 10  & 10 & 0.500 \\
DMF  & single-session-assistant & 10 & 10 & 0.900 \\
DMF  & single-session-preference & 10  & 10 & 0.600 \\
DMF  & single-session-user & 10  & 10 & 0.900 \\
DMF  & temporal-reasoning & 10  & 10 & 0.400 \\
\textbf{Mem0}  & \textbf{overall} & \textbf{60} & \textbf{60} & \textbf{0.767} \\
Mem0  & knowledge-update & 10  & 10 & 1.000 \\
Mem0  & multi-session & 10  & 10 & 0.700 \\
Mem0  & single-session-assistant & 10 & 10 & 0.600 \\
Mem0  & single-session-preference & 10  & 10 & 0.800 \\
Mem0  & single-session-user & 10  & 10 & 1.000 \\
Mem0  & temporal-reasoning & 10  & 10 & 0.500 \\
\hline
\end{tabular}
\end{table}

The meanings of the columns are described as follows:

\paragraph{Group.}
The \textit{Group} are the LoCoMo groups divided in categories, reported as follows:

\begin{itemize}
\item \textbf{knowledge-update}, whether the system can recognize that the user’s situation or personal information has changed and use the updated information.
\item \textbf{multi-session}, whether the system can aggregate or compare user information across two or more sessions.
\item \textbf{single-session-assistant}, whether the system can remember information provided by the assistant within a single session.
\item \textbf{single-session-preference}, whether the system can use user information from a single session to generate a personalized response.
\item \textbf{single-session-user}, whether the system can remember information stated by the user within a single session.
\item \textbf{temporal-reasoning}, whether the system can reason using both metadata timestamps and explicit time references in the dialogue.
\end{itemize}

\paragraph{Count.}
\textit{Count} is the number of evaluation rows in the slice.

\paragraph{Judge Count.}
\textit{Jude Count} are the rows with a judge score.

\paragraph{Avg. Judge Score.}
\textit{Avg. Judge Score} is the average of the correct answers evaluated by a LLM-judge (accuracy). The judge score is evaluated using a binary approach: the answer for the evaluation is always \textit{yes} or \textit{no}.

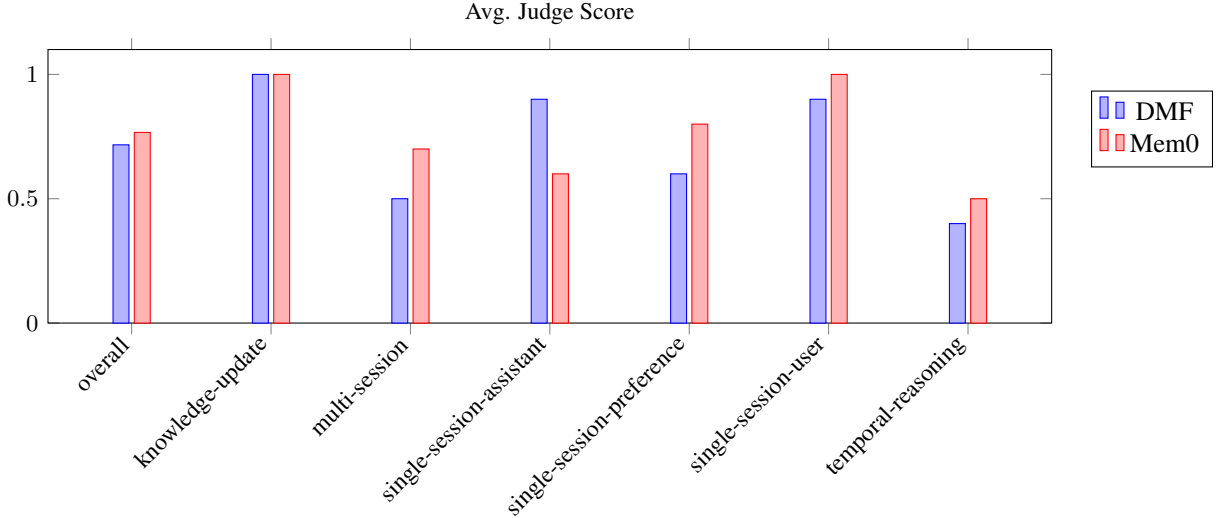
\begin{figure}[ht]
\centering
\begin{tikzpicture}

\begin{axis}[
    ybar,
    /pgf/bar width=6pt,
    width=0.9\textwidth,
    height=5.2cm,
    ymin=0,
    symbolic x coords={
        overall,
        knowledge-update,
        multi-session,
        single-session-assistant,
        single-session-preference,
        single-session-user,
        temporal-reasoning
    },
    xtick=data,
    legend style={
        at={(1.1,0.85)},
        anchor=north,
        legend columns=1
    },
    tick label style={font=\small},
    label style={font=\small},
    title style={font=\small},
    title={Avg. Judge Score},
    xticklabel style={
        rotate=45,
        anchor=east
    }
]

\addplot coordinates {
    (overall,0.7167)
    (knowledge-update,1)
    (multi-session,0.5)
    (single-session-assistant,0.9)
    (single-session-preference,0.6)
    (single-session-user,0.9)
    (temporal-reasoning,0.4)
};

\addplot coordinates {
    (overall,0.7667)
    (knowledge-update,1)
    (multi-session,0.7)
    (single-session-assistant,0.6)
    (single-session-preference,0.8)
    (single-session-user,1)
    (temporal-reasoning,0.5)
};

\legend{DMF, Mem0}

\end{axis}

\end{tikzpicture}
\caption{DMF vs. Mem0 metrics using LongMemEval-10}
\label{fig:longmemeval-metrics}
\end{figure}

The results show that DMF and Mem0 achieve similar averages, with Mem0 scoring +7\% than DMF on the Avg. Judge Score (Figure \ref{fig:longmemeval-metrics}).

For the \textbf{single-session-assistant} group, DMF outperforms Mem0 by 50\%. This is because questions in this group often require the model to recall an exact detail previously provided by the assistant, rather than a user state or preference. In this setting, DMF has an advantage because it retrieves and exposes more textual evidence, including raw turns, support records, and conversational neighbors. In contrast, Mem0 passes the output of its search process to the answerer: synthesized memories that are often user-centric and may preserve the general topic while losing numbers, lists, ordering, or exact wording.

We evaluated the token consumption of DMF and Mem0 (Table \ref{tab:longmemeval-token}\footnote{The meanings of the columns are the same of Table \ref{tab:locomo-token}.}).

On \textit{LongMemEval-10}, DMF reduces total token consumption by approximately \textbf{242×} compared with Mem0 (Rec. E2E). This reduction is primarily attributable to DMF’s deterministic memory management, which does not require LLM-based memory construction or summarization. The effect is amplified in LongMemEval, where long conversation histories make LLM-based memory processing substantially more token-intensive. As noted for LoCoMo, DMF uses a prompt with twice as many tokens as Mem0. This is necessary because DMF does not rewrite the context, as it does not use an LLM.

\begin{table}[ht]
\centering
\caption{Framework-level token usage with LongMemEval-10}
\label{tab:longmemeval-token}
\resizebox{\textwidth}{!}{%
\begin{tabular}{lrrrrrrr}
\hline
\textbf{Framework} &
\textbf{E2E Input} &
\textbf{Prompt} &
\textbf{Mem. Prompt} &
\textbf{Mem. Compl.} &
\textbf{Mem. Total} &
\textbf{Rec. E2E} \\
\hline
DMF & 569457 & 569457 & 0 & 0 & 0 &  573692 \\
Mem0 & 136536579 & 275117 & 136261462 & 2659930 & 138921392 &  139200067 \\
\hline
\end{tabular}%
}
\end{table}
\section{Conclusion}
\label{sec:conclusion}

We have presented DMF, a Deterministic Memory Framework for conversational AI agents that replaces LLM-based memory compression with a fully deterministic pipeline.
The core of DMF is a two-layer scoring model: a static Survival Score $\Omega$ combining content signals, conversational salience, and structured provenance; and a dynamic effective score $\Omega_{\mathrm{eff}}$ governed by a score-modulated exponential decay law over interaction count rather than wall-clock time.

Pruning is performed by two complementary mechanisms: a hard-kill threshold sweep that removes entries decayed below an absolute floor, and a budget-pressure procedure that computes a composite pruning score incorporating retention bonuses for operationally significant turns and penalties for superseded entries.
Long-term memory preserves raw text records as the canonical archival substrate and may add deterministic, source-linked card projections for retrieval.
This keeps the meaning of archived facts tied to stable source text while allowing query-time interpretation, filtering, and evidence assembly to evolve with the current deterministic pipeline.

The key scientific property of DMF is \emph{determinism}: for any fixed conversation sequence, the memory state is uniquely and reproducibly determined by the algorithm and configuration parameters, with no dependence on LLM sampling randomness.
This makes DMF amenable to rigorous scientific benchmarking, parameter sweeps, and audit trails.

The primary practical benefit is \emph{token efficiency}: the memory management loop uses zero LLM calls and therefore zero memory-management LLM tokens, reducing per-turn overhead by orders of magnitude compared to summarisation-based approaches.

\section{Future Work}
\label{sec:future}

Several directions for future investigation emerge naturally from the current system.

\paragraph{Multilingual support.}
The current NLP engine is English-first.
Information density and social floor heuristics depend on language-specific POS tag sets and keyword lists.
Extending DMF to additional languages requires language-specific adapters for signal extraction; the mathematical scoring and decay model are language-agnostic.

\paragraph{Adaptive scoring calibration.}
The scoring and decay parameters are currently fixed per deployment.
A future extension could use constrained reward-driven online calibration to tune selected scoring weights from retrieval success, user corrections, preference stability, and answerability.
Adaptation would remain bounded and auditable, affecting only the scoring policy while preserving source-canonical raw records.

\paragraph{Graph-enhanced lineage.}
The current lineage model supports dyadic relations (supersedes, corrects, invalidates).
A graph-based lineage structure would allow multi-hop relation chains (e.g.\ A corrects B which supersedes C) to be exploited during recall-time suppression, potentially improving precision in long-horizon correction-chain scenarios.

\paragraph{Embedding model ablation.}
DMF currently uses a single fixed embedding model.
A systematic ablation over the embedding model family (bi-encoder vs.\ cross-encoder, dimension size, training domain) would quantify the sensitivity of recall quality to the embedding choice.

\paragraph{Token budget optimisation.}
The current token budget is a static configuration parameter.
A dynamic budget policy that adjusts capacity based on conversation domain complexity or downstream task requirements could improve the trade-off between context richness and computational cost.

\paragraph{Evidence-grounded Q\&A caching.}
DMF also enables deterministic caching of recurrent questions by keying cache entries on the parsed query frame, retrieved evidence identifiers, memory snapshot, and downstream prompt/model version.
Such a cache could reuse prior answers only when their supporting raw records remain active and non-invalidated, preserving provenance and avoiding stale response reuse.

\paragraph{Shared memory.}
The source-canonical design of DMF naturally supports shared memory substrates across agents.
A DMF store can be exposed as a read-only reference memory for another agent, or multiple agents can write to a common namespace, producing an offline shared conversation mediated by deterministic retrieval rather than synchronised context windows.
Future work should study identity attribution, access control, namespace isolation, and conflict handling so that shared memories can be reused without erasing source provenance or contaminating private state.

\section*{Acknowledgments}

This work was initiated as part of the activities of the Master’s program in Data Analytics\footnote{https://master-data-analytics.it/}, organized by the Department of Industrial Engineering and the Department of Mathematics and Physics at Roma Tre University, Italy.

\printbibliography

\clearpage
\appendix
\section{Appendix}
\label{sec:appendix}

This appendix reports the prompt templates used in the benchmark pipeline to
evaluate DMF against Mem0. The templates are shared across frameworks; only the
memory context rendered into the prompt changes across systems.

\subsection*{Answer-generation prompts}

The answerer system prompt is empty in both benchmark runners. This keeps the
answer-generation phase controlled by the dataset-specific user prompt and by
the retrieved memory context.

\paragraph{LoCoMo.}
For LoCoMo, the standard prompt asks for a short answer grounded in the provided
context. Temporal questions additionally append a date-resolution instruction.

\begin{lstlisting}[language={},caption={LoCoMo answer-generation prompt.},breaklines=true,basicstyle=\ttfamily\small,columns=fullflexible]
{context}

Based on the above context, write an answer in the form of a short phrase for the following question.
Answer with exact words from the context whenever possible. Question: {question} Short answer:
\end{lstlisting}

\begin{lstlisting}[language={},caption={LoCoMo temporal-category instruction.},breaklines=true,basicstyle=\ttfamily\small,columns=fullflexible]
Use the Session Date of the conversation to answer with an approximate date.
\end{lstlisting}

\paragraph{LongMemEval.}
For LongMemEval, the answerer receives the retrieved history chats, the question
date, and the question. The rationale is to expose temporal metadata directly
without adding task-specific reasoning instructions.

\begin{lstlisting}[language={},caption={LongMemEval answer-generation prompt.},breaklines=true,basicstyle=\ttfamily\small,columns=fullflexible]
I will give you several history chats between you and a user.
Please answer the question based on the relevant chat history.


History Chats:

{context}

Current Date: {question_date}
Question: {question}
Answer:
\end{lstlisting}

\subsection*{Validation prompt}

The validation phase uses the same LLM-as-judge prompt for both datasets.
LongMemEval additionally fills the optional question type and question date
fields when they are available. The prompt favours semantic correctness over
surface-form matching, which is appropriate because the answerer is required to
produce short natural-language answers rather than exact JSON targets.

\begin{lstlisting}[language={},caption={Judge system prompt.},breaklines=true,basicstyle=\ttfamily\small,columns=fullflexible]
You are evaluating conversational AI memory recall. Return JSON only with the format requested.
\end{lstlisting}

\begin{lstlisting}[language={},caption={Judge user prompt.},breaklines=true,basicstyle=\ttfamily\small,columns=fullflexible]
Label the generated answer as CORRECT or WRONG.

{context_block}

## Rules

1. **PARTIAL CREDIT**: If the generated answer includes AT LEAST ONE correct item from the gold answer's list, mark CORRECT. Getting 1 out of 2, 2 out of 4, etc. is always acceptable. Only mark WRONG if NONE of the gold answer items appear.

2. **PARAPHRASES COUNT**: Same concept in different words is CORRECT. Judge semantic meaning, not exact wording.

3. **EXTRA DETAIL IS FINE**: A longer answer that includes the gold answer's key facts plus additional information is CORRECT. Never penalize for being more detailed or specific.

4. **DATE TOLERANCE**: Dates within 14 days of each other are CORRECT. Durations within 50% are CORRECT. Relative dates that point to the same time window are CORRECT.

5. **ABSTENTION MATCHING**: If the gold answer is an abstention or indicates the information is unavailable, any semantically equivalent refusal to answer is CORRECT.

6. **SEMANTIC OVERLAP**: Judge whether the generated answer addresses the same topic and captures the core idea of the gold answer. Different wording, phrasing, or level of detail should not result in WRONG if the underlying concept matches.

7. **SAME REFERENT**: If the generated answer identifies the same named entity, person, character, place, or concept as the gold answer, mark CORRECT even if it gives a different description or extra detail.

8. **FOCUS ON KNOWLEDGE, NOT WORDING**: The goal is to assess whether the system recalled the right fact. Minor differences in specificity, phrasing, or scope should not result in WRONG. Only mark WRONG when the generated answer demonstrates a genuinely different or incorrect understanding.

## ONLY mark WRONG if:
- The generated answer contains ZERO correct items from the gold answer
- The answer addresses a completely different topic

## Question
Question: {question}
Gold answer: {ground_truth_answer}
Generated answer: {generated_answer}

Return JSON with "reasoning" (one sentence) and "label" (CORRECT or WRONG). Do NOT include both labels.
\end{lstlisting}

\end{document}